\documentclass[11pt]{article}

\usepackage[preprint]{latex/acl}

\usepackage{times}
\usepackage{latexsym}

\usepackage[T1]{fontenc}

\usepackage[utf8]{inputenc}

\usepackage{microtype}

\usepackage{inconsolata}

\usepackage{graphicx}
\usepackage{booktabs} 
\usepackage{makecell} 
\usepackage{adjustbox} 
\usepackage[utf8]{inputenc}
\usepackage[T1]{fontenc}
\usepackage{amsmath,amssymb,amsfonts}
\usepackage{graphicx}
\usepackage{booktabs}
\usepackage{algorithm}
\usepackage{algorithmic}
\usepackage{hyperref}
\usepackage{xcolor}
\usepackage{colortbl}
\usepackage[table]{xcolor}
\usepackage{enumitem}
\usepackage{multirow}
\usepackage{subcaption}
\usepackage[most]{tcolorbox}
\usepackage[table, dvipsnames, svgnames, x11names ]{xcolor}
\usepackage[most]{tcolorbox}
\usepackage{caption}
\usepackage{dblfloatfix}

\usepackage{float}
\usepackage{booktabs}  
\usepackage{graphicx}  
\title{HiNS: Hierarchical Negative Sampling for More Comprehensive Memory Retrieval Embedding Model}

\author{
    Motong Tian$^{\diamondsuit}$\ \  
    Allen P. Wong$^{\diamondsuit}$\ \ 
    Mingjun Mao$^\spadesuit$\ \ 
    \textbf{
    Wangchunshu Zhou$^{\diamondsuit}$\thanks{Corresponding Author}~}\\
    $^\diamondsuit$OPPO \quad
    $^\spadesuit$Zhejiang University \quad
    \\
    \fontsize{10.2pt}{0.1\baselineskip}\selectfont \texttt{zhouwangchunshu@oppo.com}
}

\begin{document}
\maketitle
\begin{abstract}

Memory-augmented language agents rely on embedding models for effective memory retrieval. However, existing training data construction overlooks a critical limitation: the hierarchical difficulty of negative samples and their natural distribution in human-agent interactions. In practice, some negatives are semantically close distractors while others are trivially irrelevant, and natural dialogue exhibits structured proportions of these types. Current approaches using synthetic or uniformly sampled negatives fail to reflect this diversity, limiting embedding models' ability to learn nuanced discrimination essential for robust memory retrieval.In this work, we propose a principled data construction framework HiNS that explicitly models negative sample difficulty tiers and incorporates empirically grounded negative ratios derived from conversational data, enabling the training of embedding models with substantially improved retrieval fidelity and generalization in memory-intensive tasks.
Experiments show significant improvements: on LoCoMo, F1/BLEU-1 gains of 3.27\%/3.30\% (MemoryOS) and 1.95\%/1.78\% (Mem0); on PERSONAMEM, total score improvements of 1.19\% (MemoryOS) and 2.55\% (Mem0).

\end{abstract}

\section{Introduction}

Memory-augmented systems have become increasingly central to building intelligent agents that can learn from and reason over past interactions. A key enabler of such systems is the embedding model responsible for retrieving relevant memory entries in response to a given query or context. These models are typically trained using contrastive learning, where a positive memory is paired with one or more negative (i.e., irrelevant) memories to encourage the model to separate relevant from irrelevant information in the embedding space.

However, current approaches to constructing training data for memory retrieval largely treat all negative samples as equally uninformative—often sampling them uniformly at random or from a fixed set of hard negatives\cite{li2023towards,xiao2024c}. This overlooks two crucial aspects of real-world memory retrieval: (1) negative samples exhibit varying levels of semantic or contextual similarity to the query, forming a spectrum of difficulty that is essential for learning fine-grained discrimination\cite{ho2025arcmemo}; and (2) in natural dialogue or user interactions, different types of negatives (e.g., topic-related but factually wrong, temporally mismatched, or completely off-topic) appear in characteristic proportions that reflect human reasoning and communication patterns\cite{wang2023augmenting}. Ignoring these factors leads to embedding models that generalize poorly in realistic memory-intensive settings, where the ability to distinguish subtle distractors is paramount.




\begin{figure*}[t]
\centering
\begin{minipage}{0.88\textwidth}
\vspace{-0.1in}
    \centering
    \includegraphics[height=0.35\textheight,width=0.8\textwidth]{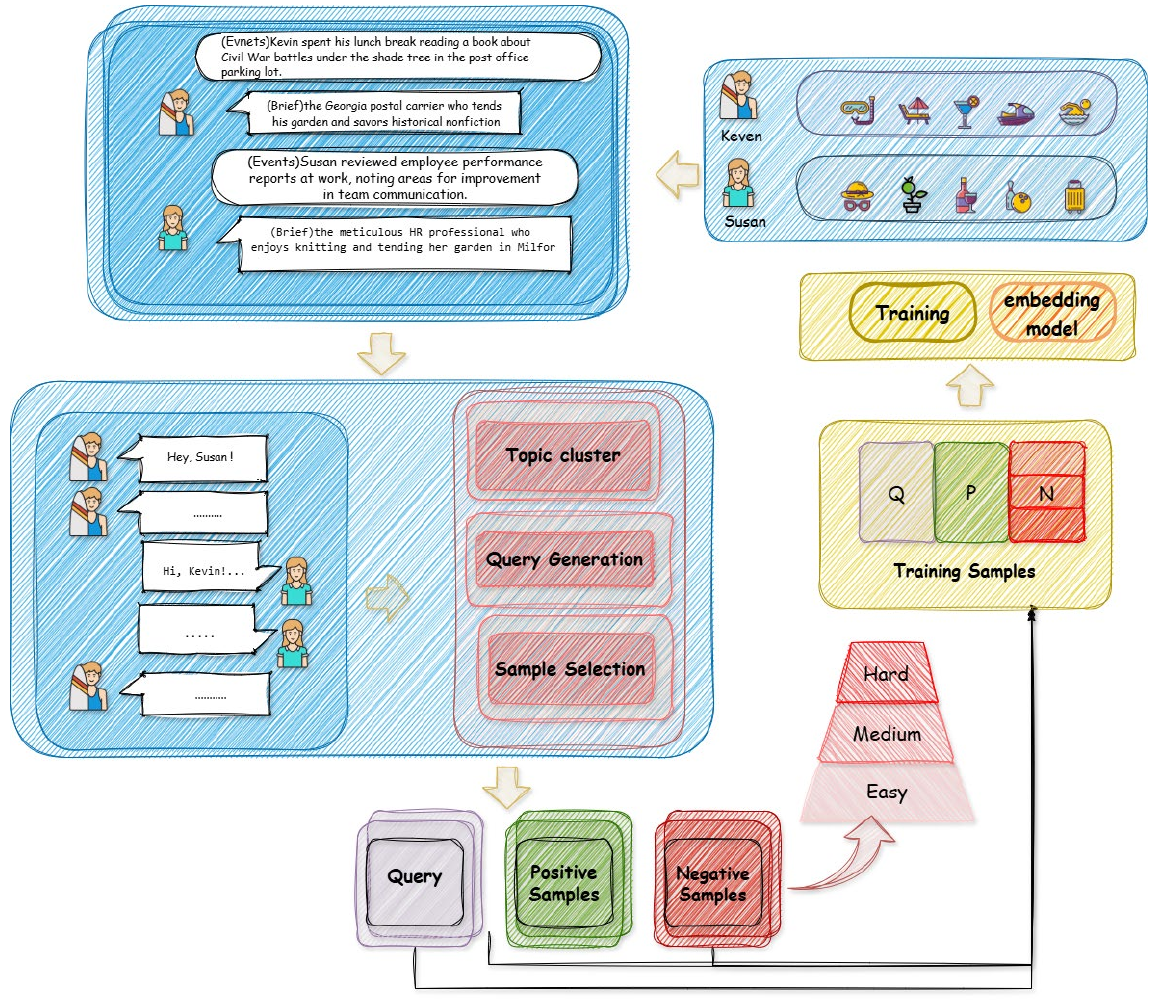}
    \captionof{figure}{\small The whole pieline of our framework.
The top-right corner presents the original persona based on Nemotron-Personas\cite{https://huggingface.co/datasets/nvidia/Nemotron-Personas}. Based on the personas, events are generated. Subsequently, natural dialogues between characters are generated according to temporal sequences, followed by topic clustering, query generation, and hierarchical negative sampling.
}
    \label{fig:main}
\end{minipage}

\end{figure*}

To address these limitations, we re-examine the data construction pipeline for memory retrieval embedding models through the lens of negative sample quality and distribution. We first identify that the neglect of hierarchical negative difficulty and the absence of dialogue-informed negative ratios in existing methods lead to \textit{poorly calibrated decision boundaries}, causing models to either overfit to trivial negatives or fail on semantically plausible distractors—both of which severely undermine retrieval robustness in real applications.

To remedy this, we propose a \textit{stratified negative sampling framework} that explicitly categorizes negatives into \textit{easy}, \textit{medium}, and \textit{hard} tiers based on contextual and semantic similarity to the query \autoref{fig:main}, and we calibrate their proportions using empirical analysis of natural human--agent conversations. This ensures that the training signal reflects both the challenge spectrum and the natural prevalence of different error types encountered in practice. Our contributions are threefold:


\begin{enumerate}
    \item \textbf{Problem identification}: We reveal that prior training data construction strategies ignore the hierarchical difficulty of negative samples and the realistic distribution of negative types in dialogue, leading to embedding models with brittle and non-discriminative memory retrieval behavior.
    \item \textbf{Methodological solution}: We introduce a principled, three-tier (easy--medium--hard) negative sampling and weighting scheme grounded in conversational data, enabling more cognitively plausible and effective contrastive learning for memory retrieval.

    \item \textbf{Empirical validation}: Extensive experiments on benchmark memory-augmented dialogue datasets demonstrate that our approach significantly improves retrieval accuracy and downstream task performance over strong baselines. On LoCoMo~\citep{maharana2024evaluating}, fine-tuning yields average F1/BLEU-1 improvements of 3.27\%/3.30\% (MemoryOS) and 1.95\%/1.78\% (Mem0). On PERSONAMEM~\citep{jiang2025know}, total score improvements reach 1.19\% (MemoryOS) and 2.55\% (Mem0).

\end{enumerate}

\section{Related Work}

\subsection{Memory-Augmented Dialogue Agents}

Recent years have witnessed significant advances in memory-augmented dialogue agents \cite{wang2025mem,liang2025towards,li2025chain,zhang2025memevolve}. \cite{hu2025memory} provide a comprehensive survey of agent memory systems, analyzing memory from the perspectives of forms (token-level, parametric, latent), functions (factual, experiential, working), and dynamics (formation, evolution, retrieval).To support personalization and consistency in long-term interactions, substantial research efforts have focused on mechanisms for memory storage, indexing, and retrieval. Representative studies include \textbf{PerLTQA}\citep{du2024perltqa}, which introduced the first QA benchmark dataset tailored for personalized long-term memory; \textbf{In Prospect and Retrospect} \citep{tan2025prospect}, which proposed a reflective memory management framework that dynamically filters outdated or conflicting memories; and \textbf{ArcMemo} \citep{ho2025arcmemo}, which emphasizes \emph{abstracted} memory representations and demonstrates that standard embedding-similarity--based retrieval becomes ineffective at higher levels of abstraction. Additional contributions include \textbf{ChatDB} \cite{hu2023chatdb}, which improves long-term consistency by decoupling user facts from dialogue history through a structured memory store; \textbf{Reflexion} \cite{shinn2023reflexion}, which incorporates a language-based reflection mechanism that enables agents to autonomously correct erroneous memories. However, existing memory-augmented systems commonly overlook the empirically observed \emph{difficulty spectrum} and distributional patterns of negative samples when training their retrieval embedding models. As highlighted by studies such as \textbf{LongMem} \cite{wang2023augmenting}, errors in real dialogues exhibit a structured difficulty distribution across easy, medium, and hard categories. This oversight results in models with insufficient discriminative power against ``hard'' distractors that are semantically similar but factually contradictory, thereby undermining their capacity to sustain consistent, personalized dialogue over extended interactions.


\subsection{Text Embedding Models}

Models such as \textbf{GTE} \cite{li2023towards} enhance generalization through instruction tuning; \textbf{BGE} \cite{xiao2024c} leverages contrastive learning and large-scale alignment to refine embedding quality; \textbf{M3-Embedding} \cite{chen2024bge} employs self-knowledge distillation to support multi-granularity, multilingual, and multifunctional embeddings; and \textbf{Qwen3 Embedding} \cite{zhang2025qwen3} builds upon the Qwen3 foundation model with a multi-stage training pipeline: it first uses LLMs to synthesize high-quality, cross-lingual, and cross-domain training pairs, then applies model merging strategies to boost robustness.

\subsubsection{Hard Negative Sampling}

Hard negative sampling, a key technique for text embedding and dense retrieval, encompasses static (predefined via data/model, e.g., \cite{he2023learning,lin2020world,ren2021rocketqav2}) and dynamic (adaptive to model state, e.g., \cite{guu2020retrieval,zhan2021optimizing}; \textbf{EDHNS} \cite{han2024efficient}) paradigms that recognize negative sample heterogeneity and use model-derived similarities for difficulty categorization to boost model discrimination.

\subsubsection{Dense Retrieval Methods}

\textbf{DPR} \cite{karpukhin2020dense} established an end-to-end dense retrieval paradigm for open-domain QA using a dual-encoder architecture; \textbf{ANCE} \cite{xiong2020approximate} enables dynamic hard negative sampling via asynchronous index updates; \textbf{RocketQA} \cite{qu2021rocketqa} improves training stability with a denoised negative selection mechanism; \textbf{SyNeg} \cite{li2024syneg} leverages LLMs to generate semantically challenging hard negatives, thereby enhancing sensitivity to subtle semantic distinctions.

However, these approaches have fundamental limitations in negative sampling design. \textbf{General-purpose embedding models} (e.g., GTE, BGE, Qwen3 Embedding) treat negatives as homogeneous noise, using heuristically fixed ratios without explicit difficulty stratification. \textbf{Dense retrieval methods} (e.g., DPR, ANCE, SyNeg) adopt the "harder is better" paradigm, ignoring the continuous difficulty spectrum and co-occurrence patterns of negatives in real interactions. \textbf{Hard negative sampling} approaches, meanwhile, define difficulty via model-derived similarities, lacking empirical grounding in real dialogue error distributions.
Collectively, these limitations hinder embedding models from disambiguating semantically proximal but factually conflicting memory distractors in long-horizon interactions, resulting in weak discriminative power, poor retrieval robustness, and compromised dialogue consistency under structured memory errors.
To address these, we propose a stratified negative sampling framework. By modeling real dialogue difficulty distribution explicitly, it enhances retrieval accuracy under complex interference and improves downstream performance.

\section{Method}


We propose HiNS, a Hierarchical Negative sampling framework for training conversational embedding models. Our approach addresses the overfitting problem in retrieval-oriented dialogue understanding through three key innovations: (1) cross-conversation negative sampling, (2) hierarchical negative stratification, and (3) semantic-aware query generation. \autoref{fig:intro} illustrates our overall pipeline. Detailed prompts for each component are provided in Appendix \ref{app:prompts}.


\begin{figure*}[!ht]
\vspace{-0.8in}
\centering
\includegraphics[height = 0.42\textwidth,width=0.9\textwidth]{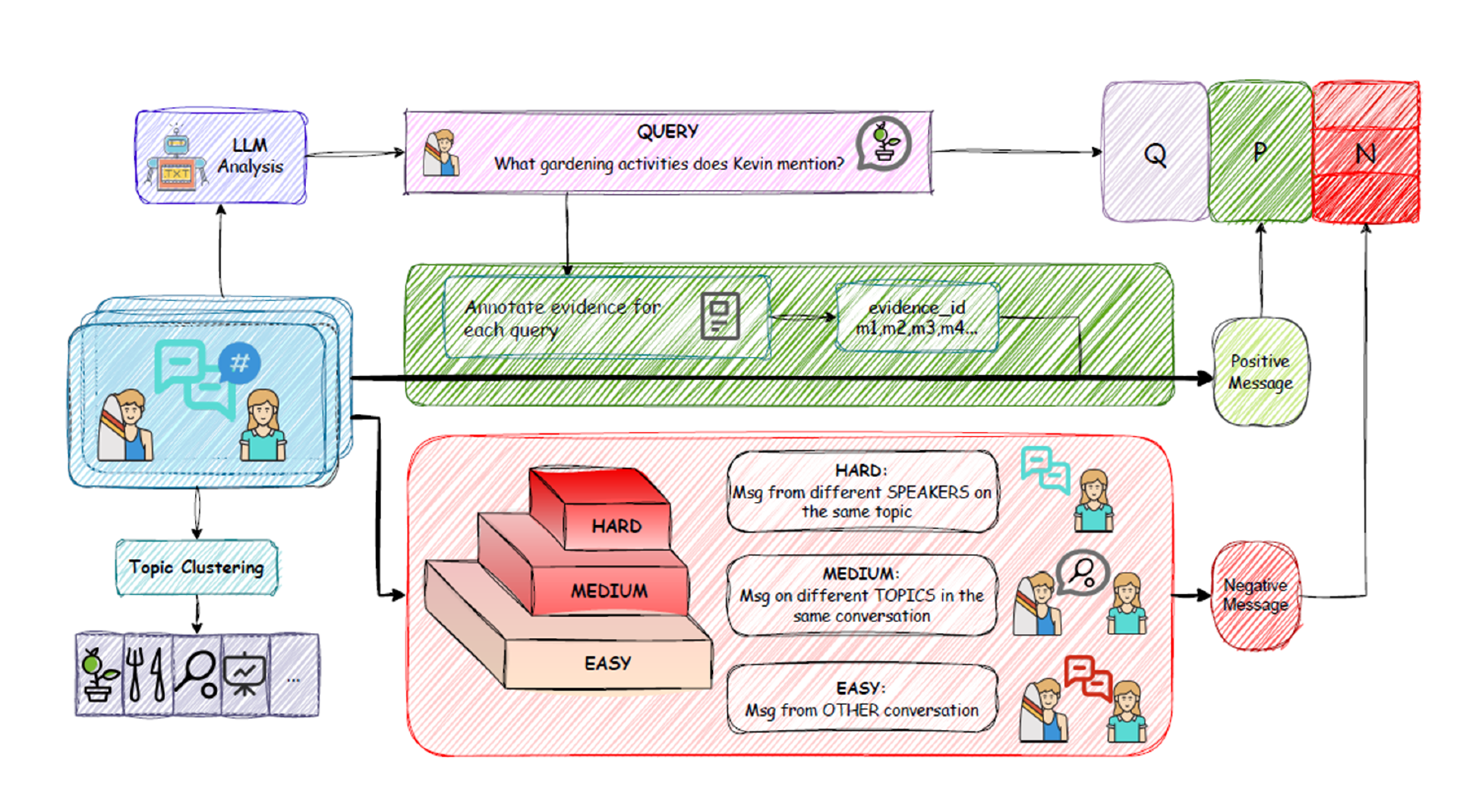}
\caption{\small
 The Hierarchical Negative Sampling process
}
\label{fig:intro}
\vspace{-0.5cm}
\end{figure*}

\subsection{Problem Formulation}

Given a multi-turn conversation $\mathcal{C} = \{m_1, m_2, \ldots, m_T\}$ between two participants $P_1$ and $P_2$, and a natural language query $q$, our goal is to learn an embedding function $f_\theta$ such that:
\begin{equation}
\text{sim}(f_\theta(q), f_\theta(m^+)) > \text{sim}(f_\theta(q), f_\theta(m^-))
\end{equation}
where $m^+$ denotes relevant messages and $m^-$ denotes irrelevant messages. The key challenge lies in constructing informative negative samples that prevent the model from learning superficial shortcuts.

\subsection{Persona-Grounded Conversation Synthesis}

To generate diverse and realistic training conversations, we design a persona-grounded synthesis pipeline.

\paragraph{Persona Sampling.}
For each conversation, we sample two distinct participants from a large-scale persona database (Nvidia Nemotron Personas). Each persona $\mathcal{P}_i$ comprises two attribute sets: (1) basic attributes $\mathcal{B}_i$ including demographics (sex, age, marital status), education (level, field), and location (city, state, country), and (2) personality attributes $\mathcal{A}_i$ including professional persona, interests (sports, arts, travel, culinary), skills, hobbies, and career goals. To ensure diversity while maintaining computational efficiency, we randomly sample $k=3$ personality attributes per persona:
\begin{equation}
\begin{aligned}
\mathcal{P}_i &= \{\mathcal{B}_i, \mathcal{A}_i\}, \\
\mathcal{A}_i &\sim \text{RandomSample}(\mathcal{A}_{\text{all}}, k=3)
\end{aligned}
\end{equation}
where $\mathcal{A}_{\text{all}}$ denotes all available personality attributes. This sampling strategy ensures each conversation involves distinct personas with varied characteristics.

\paragraph{Event Generation.}
Rather than directly exposing personality traits, we generate naturalistic daily events $\mathcal{E}_i$ that implicitly reflect the persona. For each participant $P_i$ with name $\text{name}_i$, we generate exactly 6 events using an LLM:
\begin{equation}
\mathcal{E}_i = \{\text{event}_1, \ldots, \text{event}_6\} = \text{LLM}(\mathcal{B}_i, \mathcal{A}_i, \text{name}_i)
\end{equation}
Each event $\text{event}_j$ is a natural language description of 10--25 words, covering four categories: (1) routine activities, (2) social interactions, (3) hobbies and interests, and (4) mundane daily tasks. Critically, events use the person's actual name (not generic pronouns like ``He/She'') and are not required to explicitly reference all persona attributes. We deliberately limit to 6 events to reduce information density and avoid overly structured patterns that could lead to shortcut learning.

\paragraph{Natural Conversation Generation.}
The conversation is synthesized as a 20-turn dialogue (40 messages total) between $P_1$ and $P_2$. Each message $m_t$ is structured as a tuple $(\text{speaker}, \text{message})$, where $\text{speaker} \in \{P_1, P_2\}$ and $\text{message}$ is the text content. The conversation generation follows:
\begin{equation}
\mathcal{C} = \{m_1, \ldots, m_{40}\} = \text{LLM}(\mathcal{E}_1, \mathcal{E}_2, \text{brief}_1, \text{brief}_2)
\end{equation}
where $\text{brief}_i$ is a semantic description ( ``eg. the retired aerospace engineer who loves gardening'') generated from $\mathcal{B}_i$ and $\mathcal{A}_i$. The generation process enforces natural flow by: (1) not requiring every message to reference a specific event, (2) including greetings, reactions, and transitions, and (3) varying message lengths (5--40 words). This design prevents the model from learning identity-based shortcuts while maintaining realistic dialogue structure.

\subsection{Hierarchical Negative Sampling}

We introduce a three-tier negative sampling strategy based on semantic distance to the positive samples. This hierarchy enables the model to learn fine-grained distinctions rather than relying on superficial features.


\noindent Given a query $q$ targeting person $P_t$ on topic $\tau$, we define three negative categories:
\begin{itemize}[leftmargin=*,nosep]
    \item \textbf{Hard Negatives} $\mathcal{N}_h$: Messages from the same conversation, same topic $\tau$, but different speaker $P_{\bar{t}}$
    \item \textbf{Medium Negatives} $\mathcal{N}_m$: Messages from the same conversation but different topics $\tau' \neq \tau$
    \item \textbf{Easy Negatives} $\mathcal{N}_e$: Messages from entirely different conversations
\end{itemize}

Formally, for a query $q$ with evidence set $\mathcal{V} \subset \mathcal{C}$ (positive message IDs) and target person $P_t$ on topic $\tau$, we construct negatives as follows. Let $\mathcal{M} = \mathcal{C} \setminus \mathcal{V}$ denote all non-evidence messages in the conversation.

\textbf{Hard Negatives:} We first identify candidate messages from $\mathcal{M}$ that share the same topic $\tau$ but are spoken by a different person:

\begin{equation}
\scalebox{0.9}{$\mathcal{N}_h^{\text{candidate}} = \{m \in \mathcal{M} : \text{topic}(m) = \tau, \text{speaker}(m) \neq P_t\}$}
\end{equation}

We then randomly sample from $\mathcal{N}_h^{\text{candidate}}$ with a cap of $2 \times |\mathcal{V}|$:
\begin{equation}
\scalebox{0.9}{$\mathcal{N}_h = \text{RandomSample}(\mathcal{N}_h^{\text{candidate}}, \min(|\mathcal{N}_h^{\text{candidate}}|, 2|\mathcal{V}|))$}
\end{equation}

\textbf{Medium Negatives:} From the remaining non-evidence messages that are not hard negatives:
\begin{equation}
\mathcal{N}_m^{\text{candidate}} = \mathcal{M} \setminus \mathcal{N}_h
\end{equation}
We sample up to $|\mathcal{V}|$ messages:
\begin{equation}
\scalebox{0.9}{$\mathcal{N}_m = \text{RandomSample}(\mathcal{N}_m^{\text{candidate}}, \min(|\mathcal{N}_m^{\text{candidate}}|, |\mathcal{V}|))$}
\end{equation}


\textbf{Easy Negatives:} For conversation $\mathcal{C}_i$ in batch $\mathcal{G}$, we collect messages from all other conversations:
\begin{equation}
\mathcal{N}_e^{(i)} = \bigcup_{j \neq i} \{m : m \in \mathcal{C}_j\}
\end{equation}
and sample up to $|\mathcal{V}|$ messages:
\begin{equation}
\mathcal{N}_e = \text{RandomSample}(\mathcal{N}_e^{(i)}, \min(|\mathcal{N}_e^{(i)}|, |\mathcal{V}|))
\end{equation}

\paragraph{Batch-wise Cross-Conversation Sampling.}
To efficiently construct $\mathcal{N}_e$, we process conversations in batches of size $B=4$. This batch-level processing enables efficient sharing of messages across conversations for easy negative sampling while maintaining computational efficiency.


\subsection{Semantic Query Generation}

We propose semantic query generation with three key designs.

\paragraph{Descriptive Person Reference.}
For each participant $P_i$, we generate a semantic \textit{person brief} $\text{brief}_i$ (15--25 words) that combines demographic/occupational traits with interests:
\begin{equation}
\text{brief}_i = \text{LLM}(\text{name}_i, \mathcal{B}_i, \mathcal{A}_i)
\end{equation}
Examples include ``the retired aerospace engineer who loves gardening and historical fiction'' or ``the young Atlanta-based yoga enthusiast and aspiring chef''. During query generation, the LLM receives the conversation $\mathcal{C}$ along with participant information $(P_1, \text{brief}_1)$ and $(P_2, \text{brief}_2)$, and generates queries that reference participants using their names (e.g., ``What hobbies does James mention?'') or brief-derived descriptive phrases (e.g., ``How does the retired teacher spend weekends?'').

\paragraph{Topic-Aware Query Generation.}
We generate queries through a two-stage process. First, we perform topic clustering on the conversation to identify $K$ main topics:
\begin{equation}
\mathcal{T} = \text{TopicCluster}(\mathcal{C}) = \{(\tau_k, \mathcal{M}_1^k, \mathcal{M}_2^k)\}_{k=1}^{K}
\end{equation}
where $\mathcal{M}_j^k$ denotes the set of message IDs from participant $P_j$ on topic $\tau_k$. Then, we generate exactly 6 queries $\mathcal{Q} = \{q_1, \ldots, q_6\}$ using an LLM, ensuring balanced coverage:
\begin{itemize}[leftmargin=*,nosep]
    \item 2 queries targeting $P_1$'s activities and interests
    \item 2 queries targeting $P_2$'s activities and interests  
    \item 2 queries about shared topics or interactions between both participants
\end{itemize}
Each query $q_i$ is associated with: (1) $\text{query\_text}$: the natural language query, (2) $\text{target\_person}$: $P_1$ or $P_2$, (3) $\text{topic}$: the primary topic $\tau_k$, and (4) $\text{evidence}$: a set of message IDs $\mathcal{V}_i \subset \mathcal{C}$ that directly answer the query.

\paragraph{Evidence Calibration.}
For each query, the evidence set $\mathcal{V}_i$ contains 5--12 message IDs, carefully calibrated to avoid both sparse supervision (too few positives leading to weak learning signals) and noisy labels (too many loosely related messages that dilute the positive signal). The evidence messages are primarily from the target person's messages on the relevant topic.

\subsection{Training Objective}

For each query $q$ with evidence set $\mathcal{V}$, we construct training triplets:

\begin{equation}
\begin{split}
\mathcal{D} = \{(q, m^+, \{n_1, \ldots, n_K\}) : {} \\
m^+ \in \mathcal{V}, n_k \in \mathcal{N}_h \cup \mathcal{N}_m \cup \mathcal{N}_e\}
\end{split}
\end{equation}

The final training objective follows the InfoNCE loss~\cite{oord2018representation} with hierarchical negatives.

\section{Experiments}

\subsection{Experimental Settings}

\paragraph{Dataset.}
We fine-tune our embedding model on a dataset of 201,462 training samples.

\paragraph{Model Architecture.}
We initialize our model from \texttt{BAAI/bge-small-en-v1.5}, a state-of-the-art bilingual general embedding model with 384-dimensional output embeddings. The base model employs a BERT-like architecture optimized for semantic similarity tasks.

\paragraph{Training Configuration.}
We train the model using Distributed Data Parallel (DDP) across 4 NVIDIA A800 GPUs with 80GB memory each. The training hyperparameters are as follows:
\begin{itemize}
    \item \textbf{Batch size:} 512 total (128 per GPU)
    \item \textbf{Learning rate:} $2 \times 10^{-5}$ with linear warmup over the first 10\% of training steps
    \item \textbf{Temperature parameter:} $\tau = 0.02$ for the contrastive loss
    \item \textbf{Negative sampling:} 15 negatives per query-positive pair, sampled with ratios of 30\% hard, 30\% medium, and 40\% easy negatives
    \item \textbf{Loss function:} InfoNCE loss using only explicit negatives (in-batch negatives disabled to avoid false negatives in memory retrieval scenarios)
    \item \textbf{Training duration:}final model checkpoint selected at step 490 (approximately 1.25 epochs)
\end{itemize}

\paragraph{Negative Sampling Strategy.}
To address the false negative problem inherent in in-batch negative sampling for memory retrieval tasks, we disable in-batch negatives and rely exclusively on our hierarchical explicit negatives. For each training sample, we sample 15 negatives with the following distribution: 30\% hard negatives (same conversation, same topic, different speaker), 30\% medium negatives (same conversation, different topic), and 40\% easy negatives (different conversations). This strategy ensures that all negatives are true negatives, preventing the model from learning incorrect associations.


\subsection{Baseline Comparisons}
We compare our fine-tuned model against the base \texttt{BGE-small-en-v1.5} model without fine-tuning.

We evaluate our method on two benchmarks:  
\textbf{LoCoMo}~\citep{maharana2024evaluating} emphasizes dynamic, context-aware memory retrieval across multiple real-world scenarios (e.g., temporal reasoning, open-domain dialogue, multi-hop memory association, and adversarial distractions). 
 
\textbf{PERSONAMEM}~\citep{jiang2025know} is tailored to evaluate the model’s ability to retrieve identity-consistent memories. And 
We evaluate our models using task-specific metrics: F1 and BLEU-1 scores for LoCoMo, multiple-choice accuracy for PERSONAMEM.

For the two memory benchmarks (LoCoMo and PERSONAMEM), we adopt two open-source agent memory frameworks: MemoryOS \citep{kang2025memory} and Mem0 \citep{chhikara2025mem0}. The selection of these two frameworks is motivated by their representativeness of distinct memory system design paradigms in the agent community, ensuring our evaluation covers both baseline and advanced memory architectures to comprehensively validate the model’s adaptability. Specifically:
Mem0 serves as a lightweight, baseline memory system that embodies the most fundamental memory retrieval paradigm—relying on direct cosine similarity matching between query and memory embeddings without additional semantic enhancement or structural optimization. It represents the "naive" yet widely used memory design in preliminary agent research, providing a performance lower bound for evaluating the intrinsic effectiveness of our fine-tuning strategy.
MemoryOS, by contrast, is an advanced, task-optimized memory framework tailored for personalized agent scenarios. It integrates key design innovations (e.g., hierarchical memory categorization and context-aware semantic alignment) that address core challenges in real-world memory retrieval. As such, it represents state-of-the-art memory system architectures that prioritize retrieval fidelity and adaptability to complex user interactions.
Given considerations of cost-effectiveness and accessibility, we utilize the open-source version of Mem0 for evaluations. We systematically evaluate the performance of both the original bge-small-en-v1.5 model and its fine-tuned variant across each benchmark under these two distinct memory system configurations, enabling a rigorous comparison of how fine-tuning impacts retrieval capabilities under varying memory architectural constraints.And we evaluate our model on on several general benchmark datasets.The results are in Appendix \ref{app:works}

\begin{table*}[htbp]
    \centering
    \caption{Performance comparison on LoCoMo benchmark (BGE-Small). Each cell shows F1 / BLEU-1 scores.}
    \label{tab:locomo_results_small}
    \renewcommand{\arraystretch}{1.3}
    \resizebox{\textwidth}{!}{%
    \large
    \begin{tabular}{lccccccccccc}
    \toprule
    & \textbf{Setting} & \multicolumn{2}{c}{\textbf{Temporal}} & \multicolumn{2}{c}{\textbf{Open-domain}} & \multicolumn{2}{c}{\textbf{Multi-hop}} & \multicolumn{2}{c}{\textbf{Single-hop}} & \multicolumn{2}{c}{\textbf{Average}} \\
    & & F1 & BLEU-1 & F1 & BLEU-1 & F1 & BLEU-1 & F1 & BLEU-1 & F1 & BLEU-1 \\
    \midrule
    MemoryOS & Baseline & 0.4514 & 0.3920 & 0.2834 & 0.2343 & 0.3994 & 0.3048 & 0.5170 & 0.4519 & 0.4672 & 0.3989 \\
             & \cellcolor{LightSlateBlue!30}After & \cellcolor{LightSlateBlue!30}0.4745 & \cellcolor{LightSlateBlue!30}0.4072 & \cellcolor{LightSlateBlue!30}0.3148 & \cellcolor{LightSlateBlue!30}0.2657 & \cellcolor{LightSlateBlue!30}0.4208 & \cellcolor{LightSlateBlue!30}0.3370 & \cellcolor{LightSlateBlue!30}0.5573 & \cellcolor{LightSlateBlue!30}0.4920 & \cellcolor{LightSlateBlue!30}0.4999 & \cellcolor{LightSlateBlue!30}0.4319 \\
    \cmidrule{1-12}
    Mem0 & Baseline & 0.0738 & 0.0598 & 0.1890 & 0.1388 & 0.2645 & 0.1898 & 0.2743 & 0.2358 & 0.2254 & 0.1846 \\
         & \cellcolor{LightSlateBlue!30}After & \cellcolor{LightSlateBlue!30}0.0974 & \cellcolor{LightSlateBlue!30}0.0834 & \cellcolor{LightSlateBlue!30}0.2400 & \cellcolor{LightSlateBlue!30}0.1826 & \cellcolor{LightSlateBlue!30}0.2690 & \cellcolor{LightSlateBlue!30}0.1972 & \cellcolor{LightSlateBlue!30}0.2937 & \cellcolor{LightSlateBlue!30}0.2518 & \cellcolor{LightSlateBlue!30}0.2449 & \cellcolor{LightSlateBlue!30}0.2024 \\
    \bottomrule
    \end{tabular}%
    }
    \end{table*}


\begin{table*}[htbp]
    \centering
    \caption{Performance comparison on PERSONAMEM benchmark (BGE-Small).}
    \label{tab:knowme_results}
    \renewcommand{\arraystretch}{1.3}
    \resizebox{\textwidth}{!}{%
    \large
    \begin{tabular}{lccccccccc}
    \toprule
    & \textbf{Setting} & \textbf{Pref. Rec.} & \textbf{Recall Facts} & \textbf{Suggest Ideas} & \textbf{Generalize} & \textbf{Recall Reasons} & \textbf{Track Pref.} & \textbf{Recall Mentioned} & \textbf{Total} \\
    \midrule
    MemoryOS & Baseline & 0.3818 & 0.4754 & 0.0323 & 0.3333 & 0.7980 & 0.5755 & 0.5294 & 0.4092 \\
             & \cellcolor{LightSlateBlue!30}After & \cellcolor{LightSlateBlue!30}0.4545 & \cellcolor{LightSlateBlue!30}0.4885 & \cellcolor{LightSlateBlue!30}0.0538 & \cellcolor{LightSlateBlue!30}0.3158 & \cellcolor{LightSlateBlue!30}0.7879 & \cellcolor{LightSlateBlue!30}0.5683 & \cellcolor{LightSlateBlue!30}0.5294 & \cellcolor{LightSlateBlue!30}0.4211 \\
    \cmidrule{1-10}
    Mem0 & Baseline & 0.5091 & 0.3377 & 0.1183 & 0.4912 & 0.8384 & 0.5540 & 0.2941 & 0.4669 \\
         & \cellcolor{LightSlateBlue!30}After & \cellcolor{LightSlateBlue!30}0.5091 & \cellcolor{LightSlateBlue!30}0.3311 & \cellcolor{LightSlateBlue!30}0.1613 & \cellcolor{LightSlateBlue!30}0.5088 & \cellcolor{LightSlateBlue!30}0.8687 & \cellcolor{LightSlateBlue!30}0.5612 & \cellcolor{LightSlateBlue!30}0.2941 & \cellcolor{LightSlateBlue!30}0.4924 \\
    \bottomrule
    \end{tabular}%
    }
    \vspace{1mm}
    \footnotesize
    \end{table*}

\subsection{Performance Comparison}


Table~\ref{tab:locomo_results_small} and Table~\ref{tab:knowme_results} present comprehensive performance comparisons between the baseline \texttt{BGE-small-en-v1.5} model and our fine-tuned variant across LoCoMo and PERSONAMEM benchmarks under both MemoryOS and Mem0 frameworks.

\textbf{LoCoMo Benchmark.} On the LoCoMo benchmark, our fine-tuned model demonstrates consistent improvements across all evaluated scenarios. Under the MemoryOS framework, fine-tuning yields substantial gains: average F1 score increases from 0.4672 to 0.4999 (+3.27\%) and BLEU-1 score improves from 0.3989 to 0.4319 (+3.30\%). The improvements are particularly pronounced in the Single-hop scenario, where F1 and BLEU-1 scores rise from 0.5170/0.4519 to 0.5573/0.4920 (+4.03\%/+4.01\%), and in the Open-domain setting, where F1 improves from 0.2834 to 0.3148 (+3.14\%). Under the Mem0 framework, fine-tuning still delivers meaningful improvements: average F1 increases from 0.2254 to 0.2449 (+1.95\%) and BLEU-1 from 0.1846 to 0.2024 (+1.78\%). Notably, the Open-domain scenario shows the largest absolute improvement, with F1 score increasing from 0.1890 to 0.2400 (+5.10\%).

\textbf{PERSONAMEM Benchmark.} On PERSONAMEM, fine-tuning exhibits differential effects depending on the memory framework. Under MemoryOS, the fine-tuned model achieves a total score of 0.4211, compared to 0.4092 for the baseline (+1.19\%). The most significant improvements are observed in Preference Recommendations (0.3818 $\rightarrow$ 0.4545, +7.27\%) and Suggest New Ideas (0.0323 $\rightarrow$ 0.0538, +2.15\%). Under Mem0, fine-tuning yields more substantial gains: the total score improves from 0.4669 to 0.4924 (+2.55\%). 

\textbf{Cross-Framework Analysis.} The results reveal that fine-tuning benefits both memory frameworks.This suggests that fine-tuning effectively enhances embedding quality regardless of the underlying memory system's sophistication.

\section{Discussion}

\begin{table*}[htbp]
    \centering
    \caption{Performance comparison on LoCoMo benchmark (BGE-Base, Scaling Experiment). Each cell shows F1 / BLEU-1 scores.}
    \label{tab:locomo_results_base}
    \renewcommand{\arraystretch}{1.3}
    \resizebox{\textwidth}{!}{%
    \large
    \begin{tabular}{lccccccccccc}
    \toprule
    & \textbf{Setting} & \multicolumn{2}{c}{\textbf{Temporal}} & \multicolumn{2}{c}{\textbf{Open-domain}} & \multicolumn{2}{c}{\textbf{Multi-hop}} & \multicolumn{2}{c}{\textbf{Single-hop}} & \multicolumn{2}{c}{\textbf{Average}} \\
    & & F1 & BLEU-1 & F1 & BLEU-1 & F1 & BLEU-1 & F1 & BLEU-1 & F1 & BLEU-1 \\
    \midrule
    MemoryOS & Baseline & 0.4505 & 0.3854 & 0.2393 & 0.1965 & 0.3645 & 0.2685 & 0.5173 & 0.4507 & 0.4581 & 0.3879 \\
             & \cellcolor{LightSlateBlue!30}After & \cellcolor{LightSlateBlue!30}0.4783 & \cellcolor{LightSlateBlue!30}0.4074 & \cellcolor{LightSlateBlue!30}0.2952 & \cellcolor{LightSlateBlue!30}0.2458 & \cellcolor{LightSlateBlue!30}0.4264 & \cellcolor{LightSlateBlue!30}0.3335 & \cellcolor{LightSlateBlue!30}0.5491 & \cellcolor{LightSlateBlue!30}0.4848 & \cellcolor{LightSlateBlue!30}0.4960 & \cellcolor{LightSlateBlue!30}0.4261 \\
    \cmidrule{1-12}
    Mem0 & Baseline & 0.0689 & 0.0553 & 0.1835 & 0.1351 & 0.2706 & 0.2014 & 0.2846 & 0.2440 & 0.2307 & 0.1901 \\
         & \cellcolor{LightSlateBlue!30}After & \cellcolor{LightSlateBlue!30}0.1248 & \cellcolor{LightSlateBlue!30}0.1038 & \cellcolor{LightSlateBlue!30}0.2311 & \cellcolor{LightSlateBlue!30}0.1826 & \cellcolor{LightSlateBlue!30}0.3566 & \cellcolor{LightSlateBlue!30}0.2587 & \cellcolor{LightSlateBlue!30}0.3603 & \cellcolor{LightSlateBlue!30}0.3143 & \cellcolor{LightSlateBlue!30}0.3025 & \cellcolor{LightSlateBlue!30}0.2520 \\
    \bottomrule
    \end{tabular}%
    }
    \end{table*}

\begin{table*}[t]
\centering
\small
\caption{Ablation study on LoCoMo benchmark (BGE-Small, Mem0) with different negative sampling compositions. Each cell reports F1 / BLEU-1 scores.}
\label{tab:locomo_ablation_negative_types}
\begin{tabular}{lccccc}
\hline
\textbf{Method} & \textbf{Temporal} & \textbf{Open-domain} & \textbf{Multi-hop} & \textbf{Single-hop} & \textbf{Average} \\
\hline
Just Hard (H)       & 0.0925 / 0.0768 & 0.1850 / 0.1354 & 0.2399 / 0.1712 & 0.2720 / 0.2339 & 0.2233 / 0.1836 \\
No Medium (H+E)     & 0.1070 / 0.0885 & 0.1874 / 0.1398 & 0.2678 / 0.1975 & 0.2845 / 0.2441 & 0.2384 / 0.1966 \\
No Easy (H+M)       & 0.0998 / 0.0863 & 0.1898 / 0.1388 & 0.2578 / 0.1886 & 0.2561 / 0.2183 & 0.2197 / 0.1804 \\
\rowcolor{LightSlateBlue!30}Full (E+M+H)        & 0.0974 / 0.0834 & 0.2400 / 0.1826 & 0.2690 / 0.1972 & 0.2937 / 0.2518 & 0.2449 / 0.2024 \\
\hline
\end{tabular}
\end{table*}

\subsection{Embedding Model Scaling Experiment}

In this section, we investigate whether the effectiveness of our data synthesis pipeline scales with the capacity of the underlying embedding model. While our main scaling experiment employs BGE-Small and BGE-M3 to demonstrate this trend, we also observe consistent gains at the BGE-Base level. Specifically, as shown in Table~\ref{tab:locomo_results_base}, replacing the baseline retrieval with our memory-augmented synthesis approach leads to substantial improvements across all LoCoMo subtasks—ranging from +5.6 F1 in the Open-domain setting (Mem0) to +3.8 F1 overall (MemoryOS). These results corroborate our central hypothesis: as the base embedding model becomes more powerful, our data synthesis strategy successfully leverages its enhanced representational capacity, leading to a scalable improvement in downstream performance.

\subsection{Component Analysis}

We conduct an ablation study on the LoCoMo benchmark using BGE-Small with the Mem0 architecture to assess the contribution of synthesized examples of varying difficulty levels. As shown in Table~\ref{tab:locomo_ablation_negative_types}, removing hard negatives (\textit{NoHard}) leads to only marginal gains (+0.0085 F1 on average), suggesting that hard negatives alone are insufficient to drive performance. In contrast, removing medium-difficulty examples (\textit{NoMedium}) results in the smallest performance drop, and in some cases even slight improvement over the baseline, indicating that medium examples may introduce limited signal under this model capacity. Strikingly, removing easy examples (\textit{NoEasy}) causes a noticeable degradation in overall performance ($-$0.0057 F1), particularly in the Temporal and Single-hop settings, implying that easy synthetic instances provide essential grounding for retrieval alignment. The full pipeline (\textit{E+M+H}), which combines easy, medium, and hard examples, achieves the best average performance (0.2449 F1), outperforming all ablated variants. This confirms that a balanced mixture of difficulty levels is crucial for effective data synthesis, and that easy examples---often overlooked in contrastive learning---play a non-trivial role in stabilizing retrieval behavior in low-capacity embedding models.

\section{Conclusion}

We have shown that fine-tuning embedding models with a hierarchy-aware negative sampling strategy consistently improves memory retrieval performance across diverse agent frameworks (MemoryOS and Mem0) and benchmarks (LoCoMo, PERSONAMEM). Our approach enhances both simple and sophisticated memory systems, demonstrating its compatibility with varying architectural assumptions.

\section{Limitations}
Beyond the gains discussed earlier, our work is not without limitations. First, our negative difficulty tiers rely on rule-based heuristics derived from conversation structure; a learned difficulty predictor could better capture semantic nuance. Second, training is limited to 1.25 epochs due to data constraints, potentially underutilizing the model’s capacity—longer training with larger synthesized datasets may yield further improvements. Finally, while our method improves retrieval metrics, we do not directly measure downstream task utility (e.g., agent response quality), which remains an important direction for future evaluation.

 \bibliography{latex/custom}



\appendix


\newpage
\section{Contribution}
\begin{itemize}[leftmargin=*,nosep,label={}]
    \item \textbf{Framework Design and Paper Writing}: Motong Tian, Mingjun Mao
    \item \textbf{Paper Review and Discussion}: Allen P. Wong
\end{itemize}

\section{Appendix}
\subsection{Data Privacy}
All datasets utilized in this study are publicly available conversational benchmark datasets (e.g., PERSONAMEM, LoCoMo), which have been pre-processed to remove any personally identifying information before their public release. Accordingly, our research does not involve the use of data containing personally identifiable information, and no additional privacy protection or anonymization steps were required for the data in this work.

\subsection{Potential Risk}The training quality of embedding models is highly dependent on data distribution. If the training data contains biases or noise, it can easily lead to distortion in embedding representations, which in turn undermines the reliability of downstream retrieval and matching tasks—posing a core performance risk.
Insufficient semantic alignment accuracy of embedding vectors may cause retrieval confusion risks. Meanwhile, if sensitive information embeddings are not effectively isolated during model training, there may be potential hidden dangers of privacy information leakage.


\subsection{Prompts}

\label{app:prompts}
\begin{tcolorbox}[breakable,title=Prompt for Event Generation]
You are an intelligent assistant that generates realistic daily events for a person.

Person's Name: \texttt{\{person\_name\}}
Basic Information: \texttt{\{basic\_attr\_dict\}}
Personality \& Interests: \texttt{\{persona\_attr\_dict\}}

Requirements:
\begin{enumerate}
    \item Generate exactly 6 events that reflect this person's daily life
    \item Events should be natural and varied - NOT every event needs to directly reference personality traits
    \item Include a mix of: routine activities, social interactions, hobbies, and mundane tasks
    \item Each event should use the person's NAME (not He/She pronouns)
    \item Events should be 10-25 words each
    \item Make events specific but NOT overly detailed
    \item Events should feel like real daily activities, not a checklist of personality traits
\end{enumerate}

Output format (JSON only, no explanations):
\begin{verbatim}
{
    "event1": "{person_name} 
                "did something natural "
              "and realistic",
    "event2": "{person_name} 
            "engaged in another "
              "activity",
    "event3": "{person_name} ...",
    "event4": "{person_name} ...",
    "event5": "{person_name} ...",
    "event6": "{person_name} ..."
}
\end{verbatim}

Generate 6 varied, realistic events:
\end{tcolorbox}

\begin{tcolorbox}[breakable,title=Prompt for Natural Conversation Generation]
You are a dialogue generation expert creating natural conversations between two friends.

Person 1: \texttt{\{person1\_name\}}
\begin{itemize}
    \item Background: \texttt{\{person1\_brief\}}
    \item Recent activities: \texttt{\{person1\_events\}}
\end{itemize}

Person 2: \texttt{\{person2\_name\}}
\begin{itemize}
    \item Background: \texttt{\{person2\_brief\}}
    \item Recent activities: \texttt{\{person2\_events\}}
\end{itemize}

Generate a natural 20-turn conversation (40 messages total) with these requirements:

\begin{enumerate}
    \item \textbf{NATURAL FLOW}: Not every message needs to reference a specific event. Include:
    \begin{itemize}
        \item Greetings and small talk
        \item Follow-up questions and reactions (``Really?'', ``That sounds fun!'')
        \item Transitions between topics
        \item Emotional responses and opinions
    \end{itemize}
    
    \item \textbf{REALISTIC COVERAGE}: Over the full conversation, naturally incorporate MOST (not necessarily all) events from both people
    
    \item \textbf{VARIED MESSAGE TYPES}:
    \begin{itemize}
        \item Some messages should be short reactions (5-15 words)
        \item Some should be detailed sharing (20-40 words)
        \item Include questions, statements, and exclamations
    \end{itemize}
    
    \item \textbf{NO RIGID PATTERNS}: 
    \begin{itemize}
        \item Don't force every turn to introduce a new event
        \item Let conversations naturally drift and return to topics
        \item Some messages can be pure social interaction without event content
    \end{itemize}
    
    \item \textbf{USE NAMES NATURALLY}: Occasionally use each other's names in conversation
\end{enumerate}

Output format (JSON only):
\begin{verbatim}
{
    "m1": {
        "speaker": "{person1_name}",
        "message": "greeting or 
        opening message"
    },
    "m2": {
        "speaker": "{person2_name}",
        "message": "response"
    },
    ... continue to m40 ...
    "m40": {
        "speaker": "{person2_name}",
        "message": "closing or 
        final response"
    }
}
\end{verbatim}

\textbf{IMPORTANT}: Generate exactly 40 messages (m1 to m40). Output only valid JSON.
\end{tcolorbox}

\begin{tcolorbox}[breakable,title=Prompt for Semantic Query Generation]
You are creating search queries for a conversation retrieval system.

Conversation participants:
\begin{itemize}
    \item \texttt{\{person1\_name\}}: \texttt{\{person1\_brief\}}
    \item \texttt{\{person2\_name\}}: \texttt{\{person2\_brief\}}
\end{itemize}

Conversation:
\texttt{\{conversation\}}

Generate 6 diverse queries with these requirements:

\begin{enumerate}
    \item \textbf{SEMANTIC IDENTIFICATION}: Use descriptive phrases instead of generic labels:
    \begin{itemize}
        \item GOOD: ``What hobbies does \texttt{\{person1\_name\}} mention?''
        \item GOOD: ``How does the retired teacher spend weekends?''
        \item BAD: ``What does user1 like?'' (too generic)
    \end{itemize}
    
    \item \textbf{QUERY DIVERSITY}: Create different types:
    \begin{itemize}
        \item 2 queries about \texttt{\{person1\_name\}}'s activities/interests
        \item 2 queries about \texttt{\{person2\_name\}}'s activities/interests
        \item 2 queries about shared topics or interactions between them
    \end{itemize}
    
    \item \textbf{EVIDENCE BALANCE}: Each query should have 5-12 relevant messages (not too few, not too many)
    
    \item \textbf{SPECIFICITY}: Queries should be specific enough to have clear positive/negative distinctions
\end{enumerate}

Output format (JSON only):
\begin{verbatim}
{
    "query1": {
        "query_text": "What outdoor 
        activities does "
        "{person1_name} enjoy?",
        "target_person": 
        "{person1_name}"
        
        ,
        "topic": "outdoor activities",
        "evidence": ["m3", "m7", 
        "m15", "m23", "m31"]
    },
    "query2": {
        "query_text": "How does 
        {person2_name} describe "
                      "their work 
                      experiences?",
        "target_person": 
        "{person2_name}",
        "topic": "work",
        "evidence": ["m4", "m8", 
        "m12", "m20"]
    },
    ... continue for all 6 queries ...
}
\end{verbatim}

Requirements for evidence:
\begin{itemize}
    \item Only include message IDs where the content DIRECTLY answers the query
    \item Evidence should be from the target person's messages primarily
    \item Include 5-12 message IDs per query
    \item Be precise - not every mention of a topic counts as evidence
\end{itemize}

Generate 6 queries now:
\end{tcolorbox}

\begin{tcolorbox}[breakable,title=Prompt for Topic Clustering]
Analyze this conversation and cluster messages by topic.

Conversation:
\texttt{\{conversation\}}

Identify 4-6 main topics discussed and assign each message to its primary topic.

Output format (JSON only):
\begin{verbatim}
{
    "topics": {
        "topic1_name": {
            "description": "brief 
            description of topic",
        "messages_from_{person1_name}": 
                ["m1", "m5", "m13"],
        "messages_from_{person2_name}": 
                ["m2", "m6", "m14"]
        },
        "topic2_name": {
            "description": 
            "brief description",
        "messages_from_{person1_name}": 
                ["m3", "m9"],
        "messages_from_{person2_name}": 
                ["m4", "m10"]
        }
    }
}
\end{verbatim}

Be thorough - assign every message to exactly one topic.
\end{tcolorbox}

\begin{tcolorbox}[breakable,title=Prompt for Person Brief Generation]
Create a brief, memorable description (15-25 words) for this person that captures their key characteristics.

Name: \texttt{\{name\}}
Basic info: \texttt{\{basic\_attr\_dict\}}
Personality: \texttt{\{persona\_attr\_dict\}}

The description should:
\begin{enumerate}
    \item Be unique and identifiable
    \item Mention 2-3 key traits (age/occupation + 1-2 interests)
    \item Be natural to use in a question (e.g., ``What does [description] think about\ldots'')
\end{enumerate}

Output only the description, no quotes or extra text.

Example outputs:
\begin{itemize}
    \item ``the retired aerospace engineer who loves gardening and historical fiction''
    \item ``the young Atlanta-based yoga enthusiast and aspiring chef''
    \item ``the married software developer passionate about basketball and travel''
\end{itemize}

Generate the description:
\end{tcolorbox}

\begin{tcolorbox}[breakable,title=Prompt for Distractor Events Generation]
Given these events from one person:
\texttt{\{target\_events\}}

Generate 6 SIMILAR but DIFFERENT events that could belong to a DIFFERENT person with related interests.

Requirements:
\begin{enumerate}
    \item Events should be in the same DOMAIN (e.g., if original is about cooking, create different cooking events)
    \item But with DIFFERENT specifics (different dishes, different contexts)
    \item Use a DIFFERENT name for this hypothetical person
    \item These should be plausible ``hard negatives'' - similar enough to be confusing but clearly from a different person
\end{enumerate}

Persona type to match: \texttt{\{persona\_type\}}

Output format (JSON only):
\begin{verbatim}
{
    "distractor_name": "A 
    different person's name",
    "events": {
        "d1": "event similar to 
        original but "
              "different",
        "d2": "another similar
        but different "
              "event",
        ...
    }
}
\end{verbatim}
\end{tcolorbox}

\begin{tcolorbox}[breakable,title=Prompt for Cross-Conversation Query Generation]
You have two separate conversations:

Conversation A (between \texttt{\{person1\_name\}} and \texttt{\{person2\_name\}}):
\texttt{\{conv1\_summary\}}

Conversation B (between \texttt{\{person3\_name\}} and \texttt{\{person4\_name\}}):
\texttt{\{conv2\_summary\}}

Generate 4 queries where:
\begin{itemize}
    \item The CORRECT answer comes from Conversation A
    \item Conversation B contains SIMILAR but WRONG content (hard negatives)
\end{itemize}

Each query should:
\begin{enumerate}
    \item Ask about a specific topic that appears in BOTH conversations
    \item But with details that only match Conversation A
    \item Be specific enough that only one conversation truly answers it
\end{enumerate}

Output format (JSON only):
\begin{verbatim}
{
    "query1": {
        "query_text": "specific 
        question about topic X",
        "correct_conversation": "A",
        "correct_messages": ["m5",
        "m12", "m23"],
        "hard_negative_messages_from_B": 
            ["m3", "m8", "m15"],
        "reasoning": "why B messages 
        are similar but wrong"
    },
    ...
}
\end{verbatim}
\end{tcolorbox}

\subsection{General embedding benchmark Comparison}
\label{app:works}
To comprehensively evaluate the generalization capability of our proposed model, we perform extensive experiments on several well-recognized benchmark datasets. Our evaluation focuses on general-domain tasks, which examine the model’s core embedding quality across diverse textual contexts. Notably, the selected general-domain tasks have been widely adopted as standard evaluation benchmarks in prior text embedding research, including representative works such as M3-Embedding \cite{chen2024bge} and Qwen3 Embedding \cite{zhang2025qwen3}. This task selection enables direct and fair comparison with these state-of-the-art models, thereby facilitating rigorous validation of our model’s performance advantages.

\label{sec:appendix}
\newpage
\vspace*{-1cm}  
\noindent  
\begin{table*}[!t]  
    \centering
    \caption{General embedding benchmark comparison: Baseline(BGE-Small) vs After fine tuning .}
    \label{tab:mteb_comparison}
    \renewcommand{\arraystretch}{1.2}
    \resizebox{\textwidth}{!}{%
    \small
    \begin{tabular}{lccccl}
    \toprule
    \textbf{Task Name} & \textbf{Baseline} & \textbf{After} & \textbf{Diff.} & \textbf{Diff. (\%)} & \textbf{Status} \\
    \midrule
    TwentyNewsgroupsClustering & 0.349190 & 0.369433 & +0.020243 & +5.80\% & \checkmark \\
    ArXivHierarchicalClusteringS2S & 0.532609 & 0.552233 & +0.019624 & +3.68\% & \checkmark \\
    StackExchangeClusteringP2P.v2 & 0.383960 & 0.385161 & +0.001201 & +0.31\% & \checkmark \\
    STS12 & 0.658655 & 0.655377 & -0.003278 & -0.50\% & $\times$ \\
    ImdbClassification & 0.878528 & 0.862592 & -0.015936 & -1.81\% & $\times$ \\
    \bottomrule
    \end{tabular}%
    }
\end{table*}

\end{document}